\documentclass{article}

\usepackage{arxiv}

\usepackage[utf8]{inputenc} 
\usepackage[T1]{fontenc}    
\usepackage{hyperref}       
\usepackage{url}            
\usepackage{booktabs}       
\usepackage{amsfonts}       
\usepackage{nicefrac}       
\usepackage{microtype}      
\usepackage{lipsum}		
\usepackage{graphicx}
\usepackage{natbib}
\usepackage{doi}
\usepackage{verbatim}
\usepackage{subcaption}
\usepackage[misc]{ifsym}

\usepackage{algorithm2e}
\usepackage{amsmath}
\usepackage{array}
\usepackage{multirow}
\usepackage{subcaption}
\usepackage{listings}
\usepackage{float}

\usepackage{pgf}
\usepackage{tikz}
\usetikzlibrary{arrows,automata}
\usetikzlibrary{positioning}

\tikzset{
    state/.style={
           rectangle,
           rounded corners,
           draw=black, very thick,
           minimum height=2em,
           minimum width=2em,
           inner sep=2pt,
           text centered,
           },
}
\usepackage{placeins}
\usepackage{amssymb}
\usepackage{enumitem}
\usepackage{pifont}

\title{Automating the Analysis of Institutional Design in International Agreements}

\author{ 
\href{https://orcid.org/0000-0002-3407-7570}{\includegraphics[scale=0.06]{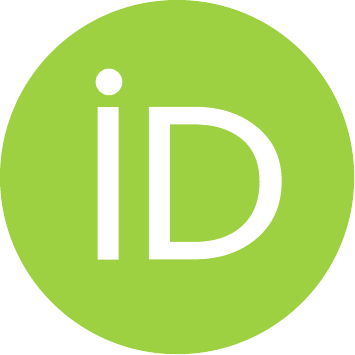}\hspace{1mm}Anna Wr{\'o}blewska$^1$\Letter},
\href{https://orcid.org/0000-0003-2664-2135}{\includegraphics[scale=0.06]{orcid.pdf}\hspace{1mm}Bartosz Pieli{\'n}ski$^2$\Letter},
\href{https://orcid.org/0000-0003-0617-7301}{\includegraphics[scale=0.06]{orcid.pdf}\hspace{1mm}Karolina Seweryn$^{1,3}$},
\href{https://orcid.org/0000-0001-5960-8131}{\includegraphics[scale=0.06]{orcid.pdf}\hspace{1mm}\textbf{Sylwia Sysko-Romańczuk}$^4$},\\
\href{https://orcid.org/0000-0002-8809-5248}{\includegraphics[scale=0.06]{orcid.pdf}\hspace{1mm}\textbf{Karol Saputa}$^1$},
 \textbf{Aleksandra Wichrowska}$^1$,
\href{https://orcid.org/0000-0002-6586-8455}{\includegraphics[scale=0.06]{orcid.pdf}\hspace{1mm}\textbf{Hanna Schreiber}$^2$},\\
 	$^1$Faculty of Mathematics and Information Science, Warsaw University of Technology, Warsaw, Poland \\ Email: \texttt{anna.wroblewska1@pw.edu.pl} \\
	$^2$Faculty of Political Science and International Studies, University of Warsaw, Warsaw, Poland \\
 Email: \texttt{b.pielinski@uw.edu.pl} \\
    $^3$NASK - National Research Institute, Warsaw, Poland \\
 Email: \texttt{karolina.seweryn@nask.pl} \\
	$^4$Faculty of Management, Warsaw University of Technology, Warsaw, Poland
 }

\date{}

\renewcommand{\shorttitle}{Entity Graph Extraction from  
Legal Acts}

\hypersetup{
pdftitle={Entity Graph Extraction from  
Legal Acts -- a Prototype for a Use Case in Policy Design Analysis},
pdfsubject={cs.LG},
pdfauthor={Anna Wróblewska, Bartosz Pieliński, Karolina Seweryn, Karol Saputa, Aleksandra Wichrowska, Sylwia Sysko-Romańczuk, Hanna Schreiber},
pdfkeywords={Natural Language Processing, Information Extraction, Policy Design, Institutional Grammar, Entity Graph, UNESCO},
}

\begin{document}
\maketitle

\begin{abstract}
This paper explores the automatic knowledge extraction of formal institutional design - norms, rules, and actors - from international agreements. The focus was to analyze the relationship between the visibility and centrality of actors in the formal institutional design in regulating critical aspects of cultural heritage relations. The developed tool utilizes techniques such as collecting legal documents, annotating them with Institutional Grammar, and using graph analysis to explore the formal institutional design. The system was tested against the 2003 UNESCO Convention for the Safeguarding of the Intangible Cultural Heritage. 
\end{abstract}

\keywords{Natural Language Processing \and Information Extraction \and Institutional Design \and Institutional Grammar \and Entity Graph \and UNESCO \and Intangible Cultural Heritage
}

\section{Introduction}
\label{sec:intro}

Solving contemporary interlinked and complex problems requires international cooperation. This cooperation is expressed through international agreements establishing norms, rules, and actors to facilitate collaboration between nations. Though digitization processes have facilitated access to official documents, the sheer volume of international agreements makes it challenging to keep up with the number of changes and understand their implications. The formal institutional design expressed in legal texts is not easily comprehensible. This is where computational diplomacy with NLP models comes into play. They hold the potential to analyze vast amounts of text effectively, but the challenge lies in annotating legal language~\cite{Libal_Tomer}. Our paper and accompanying prototype tool aims to facilitate the analysis of institutional design in international agreements. 


The authors annotated the legal text using Institutional Grammar (IG), a widely used method for extracting information from written documents~\cite{ig_codebook}. IG offers a comprehensive analysis of the rules that regulate a particular policy, revealing the relationships between rules, reducing complexity, and identifying key actors involved in the policy. Despite being relatively new, IG has achieved high standardization~\cite{ig_codebook} through close collaboration between political/international relations and computer scientists. As such, the prototype aligns well with the intersection of these two areas of scientific research. 

The growing popularity of IG in public policy studies has led to the development of several tools helping to upscale the usage of IG. There is the IG Parser created by Christopher Frantz\footnote{\url{https://s.ntnu.no/ig-parser-visual}}
, the INA Editor created by Amineh Ghorbani (\url{https://ina-editor.tpm.tudelft.nl/}), and attempts have been made to develop software for the automatization of IG annotations ~\cite{Rice2021}. The first tool is handy in manually annotating legal texts, and the second is beneficial in semi-manually transforming IG data into a network representation of rules. The third one shows the potential of IG parsing automatization. However, each tool focuses only on one aspect of the multilevel process of transforming a large corpus of legal text into a graph. The tool presented in this paper is more than a single element -- it chains together three separate modules that automate the whole process of working with IG. Although IG allows relatively easy annotation of legal text, the data produced at the end is raw. It provides no information on institutional design expressed in legal documents. This is where graphs come to the rescue.


Graphs are the most popular way of expressing the composition of rules identified by IG. Several published studies have already proved the utility of such an approach \cite{Heikkila2018,Olivier-Schlager,MESDAGHI2022120}. It makes it possible to look at elements of rules (e.g., addressees of rules or targets of action regulated by rules) as nodes connected by an edge representing the rule. If an individual rule can be modeled as an edge, then the whole regulation could be seen as a graph or a network. This conceptualization allows the introduction of measures used in network analysis to study specific policy design, making the entire field of institutional design more open to quantitative methods and analysis.


We use a graph representation of a formal institutional design to describe the position of actors in a given institutional arrangement. To achieve this aim, we use conventional network metrics associated with centrality. Using the UNESCO Convention for the
Safeguarding of the Intangible Cultural Heritage (the 2003 UNESCO Convention)
as our use case, we focus on the actors’ location in the network. We wish to address a research issue discussed in the International Public Administration (IPA) literature: the role of people working in international organizations versus states and other institutional institutions actors~\cite{eckhard_international_2016}. Researchers are interested in the degree that international civil servants are autonomous in their decisions and to what extent they can shape the agenda of international organizations. Particularly interesting is the position of treaty secretariats -- bodies created to manage administrative issues related to a specific treaty~\cite{bauer_bureaucratic_2016,jorgens_exploring_2016}. We wanted to learn about the Secretariat's position concerning other actors described by the Convention, both in data coming from IG and the Convention's network representation. 

Parsing legal texts by way of IG requires our tool to consist of three technical components: (1) a scraper, (2) a tagger, and (3) a graph modeler. Thanks to this, our prototype can scrape many legal documents from the Internet. Then, it can use the IG method, NLP techniques, and IG-labelled legal regulations to generate hypergraph representation and analyze inter-relations between crucial actors in formal institutional design. 
The aim is to use IG as an intermediate layer that makes it possible to transform a legal text into a graph which is then analyzed to learn about the characteristics of the institutional design expressed in the text. A rule, or to use IG terminology, an institutional statement, is represented here as an edge connecting actors and objects mentioned in the statement. This approach makes it possible to model a legal regulation text as a graph and analyze institutional design through indicators used in network analysis. The approach has already been implemented in research papers~\cite{Olivier2019,Olivier-Schlager,MESDAGHI2022120,unpub_Ghorbani2022}.

Our purpose while designing this prototype was to build an effective system based on existing technologies. 
All of the system's sub-modules are derived from well-known techniques (scraping, OCR, text parsing, graph building); what is innovative is how they are adjusted and coupled.\footnote{The source code of our analysis: \url{https://github.com/institutional-grammar-pl}.} To our knowledge, no other system takes input regulations distributed throughout the Internet and produces an output graph representation of institutional design. The biggest challenge in the prototype design was adjusting NLP methods for the specificity of IG. Thus, in the following use case, we concentrate on this challenge. We show how to build a graph of inter-relations between important objects and actors based on information extracted using IG from a single but crucial document.


For this paper, our prototype was tested on the 2003 UNESCO Convention. This document was chosen for three reasons. Firstly, the Convention text is relatively short and well-written, which allows for the first test to be one where the IG tagger works on a document that does not breed unnecessary confusion on the syntactic and semantic levels. Secondly, we have involved an expert on the Convention in our project, allowing us to check our analysis against her expertise.\footnote{The expert is Hanna Schreiber.} Thirdly, the Convention is considered to be one of the most successful international documents, almost universally ratified (181 States Parties as of April 2023). The document is analyzed to compare the position of institutional actors mentioned in the document.  


The following sections present our current achievements leading to our research goal. Section~\ref{sec:assumptions} highlights the current achievements in working with the IG method, developed independently, and sketches out our current research ideas. We then describe our prototype and its main modules (Section~\ref{sec:system}). Our use case -- modeling the 2003 UNESCO Convention -- is presented in Section~\ref{sec:use_case}. The paper concludes with Section~\ref{sec:conclusions}.


\section{Institutional Grammar and Graphs}
\label{sec:assumptions}

The most crucial part of our system is IG, employed as an interlayer between the regulation text and its graph representation. 

IG was used as the layer because it was designed as a schema that standardizes and organizes information on statements coming from a legal text. The statements are understood in IG as bits of institutional information. They have two main functions: to set up prominent institutional actors (organizations, collective bodies, organizational roles, etc.) and to describe what actions those actors are expected, allowed, or forbidden to perform. Therefore, IG makes it possible: (1) to identify how many statements are written into a sentence; (2) to categorize those statements; (3) to identify links between them; (4) to identify animated actors and inanimate objects regulated by the statements; and (5) to identify relations between actors and objects defined by the statements.    

\begin{table}[!htb]
 \centering
    \caption{IG components depending on statement type 
    based on~\cite[pp.~10-11]{ig_codebook}.\label{tab:ig_components}}
    \begin{tabular}{ p{2cm}|p{4cm}|p{2cm}|p{3.8cm} }
    \textbf{Regulative} & \textbf{Description} &  \textbf{Constitutive} & \textbf{Description}\\
    \hline \hline
    Attribute (A) & The addressee of the statement. &  Constituted Entity (E) & The entity being defined.\\
    \hline
    Aim (I) & The action of the addressee regulated by the statement. & Constitutive Function (F) & A verb used to define (E). \\
    \hline
    Deontic (D) & An operator determines the discretion or constraint associated with (I). & Modal (M) & An operator determining the level of necessity and possibility of defining (E).\\
    \hline
    Object (B) & The action receiver described by (I). There are two possible receivers: Direct or Indirect.  & Constituting Properties (P) & The entity against which (E) is defined.\\
    \hline
    Activation Condition (AC) & The setting to which the statement applies. & Activation Condition (AC) & The setting to which the statement applies.\\
    \hline
    Execution Constraint (EC) & Quality of action described by (I) & Execution Constraint (EC) & Quality of (F).\\
    \end{tabular}
\end{table}

\begin{figure}
\centering
\begin{subfigure}{.48\textwidth}
  \centering
    \includegraphics[width=5.5cm]{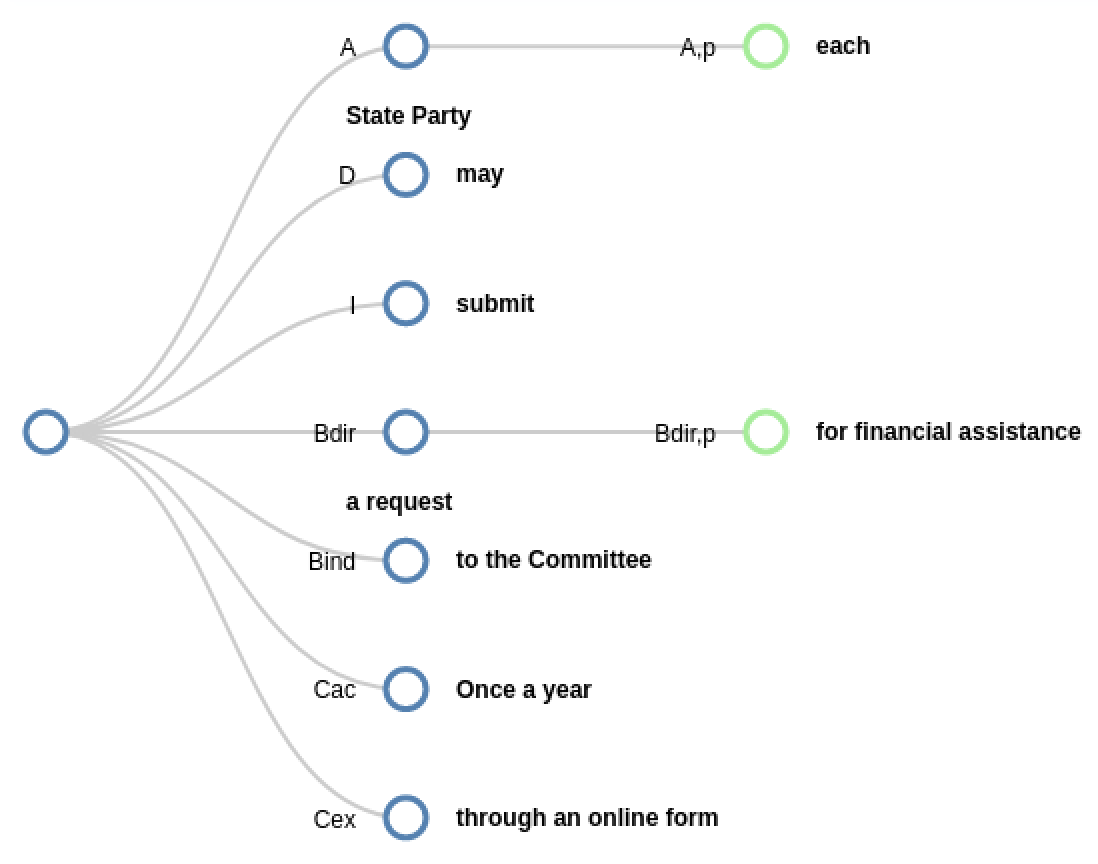}
    \caption{A regulative statement: "Once a year, each State Party may submit a request for financial assistance to the Committee through an online form"}%
    \label{fig:regulative-example}%
\end{subfigure}%
   \hspace{3mm}
\begin{subfigure}{.48\textwidth}
  \centering
    \includegraphics[width=5.5cm]{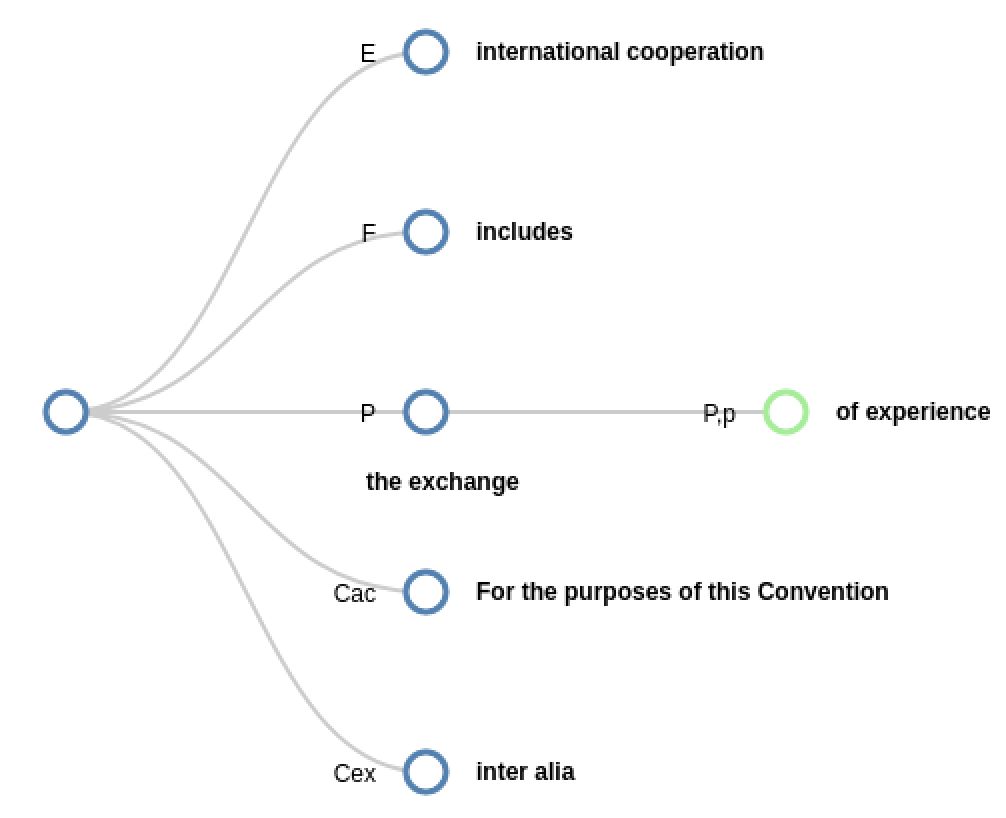}
    \caption{A constitutive statement: "For the purpose of this Convention, international cooperation includes, inter alia, the exchange of experience"}%
    \label{fig:constitutive-example}%
\end{subfigure}
\caption{Examples of parsed statements. The IG Parser generated the tree (\url{https://s.ntnu.no/ig-parser-visual}).}
\label{fig:statement-examples}
\end{figure}

Figure~\ref{fig:regulative-example} presents an example of a regulative statement from the 2003 UNESCO Convention (Article 23 par. 1). In this statement, "State Party" is an entity whose actions are regulated; therefore, it is annotated as Attribute. An action regulated explicitly in the statement is "submit," and this element is identified as Aim. The statement informs us through the Deontic "may" that the action "submit" is allowed to be performed by a State Party. However, the statement also tells through the Execution Constraint what kind of action of submission is allowed -- "through an online form." From the Activation Condition "once a year," we also learn when a State Party is entitled to the action "submit." Another piece of information provided by the statement is the kind of target directly affected by the regulated activity -- it is a Direct Object "request" that has the property "for financial assistance." The statement also identifies who is  obliquely affected by the State Party's action -- this is the Indirect Object of "the Committee."  

Another statement, a part of Article 19 par. 1, from the 2003 UNESCO Convention, is a simple example of a constitutive statement (Figure~\ref{fig:constitutive-example}). Here, the statement defines one of the essential properties of "international cooperation" that is indicated as a Constituted Entity. The statement informs that international cooperation "includes" (Function) a specific type of exchange (Constituting Property) -- the exchange "of experience" (Constituting Entity property). We also learn that this definition of international cooperation applies only in a specific context -- inside the institutional setting described by the 2003 UNESCO Convention. The Activation Condition provides information about the context in which the definition applies: "for the purposes of this Convention."

 
The application of IG makes it possible to extract standardized data on rules from legal texts. However, the IG schema is not associated with any popular way of data analysis, which is challenging to utilize in research. Therefore, researchers have been searching for ways to transform IG data into more popular data formats. The most promising direction is graph theory. Statements can be seen as edges that connect nodes represented by Attributes, Objects, Constituted Entities, and Constituting Properties. In this regard, a set of edges --statements constitutes a graph representing an individual regulation. This approach allows for using popular network analysis indicators to characterize the institutional setting developed by the regulation. A legal act is a list prescribed 
by the legal relations between different objects and actors. Three approaches have been developed to transform IG data into graph data. The first is represented by T. Heikkila and C. Weible~\cite{Heikkila2018}. IG is used here to identify actors in a variety of legal documents. They are linked in a single document and across many regulations. The aim is to identify a network of influential actors in a particular policy and use network analysis to characterize their associations. Heikkila and Weible used their method to study the regulation of shale gas in Colorado and found that the institutional actors governing this industry create a polycentric system. The second approach to IG and network analysis is represented by research by O. Thomas and E. Schlager~\cite{Olivier-Schlager,Olivier2019}. Here, data transformation from IG to graph allows for comparing the intensity with which different aspects of institutional actors' cooperation are regulated. The authors used this approach to study water management systems in the USA to link IG data with graph representation. It is built around assumptions that network analysis based on IG makes it possible to describe institutional processes written into legal documents in great detail. This approach was also used in a study on the climate adaptation of transport infrastructures in the Port of Rotterdam~\cite{MESDAGHI2022120}.  

In contrast, our approach 
is more actor-centered than the previous ones (see Section~\ref{sec:use_case} for more details). As stated in~\cite{Entity-Mention-Aware-Document-Representation}, entity mentions may provide more information than plain language in some texts. We believe that actors are crucial in analyzing the legal acts in our use case. We aim to compare the positions of prominent actors in a network of rules and use indicators developed by network analysis to describe the place of institutional actors in the formal design of the UNESCO Convention. We also confront these metrics with IG data to determine if an actor's position in a rules network correlates with its place in IG statements.  

Information structured in a graph, with nodes representing entities and edges representing relationships between entities, can be built manually or using automatic information extraction methods. Statistical models can be used to expand and complete a knowledge graph by inferring missing information~\cite{Nickel2016}.
There are different technologies and applications of language understanding, knowledge acquisition, and intelligent services in the context of graph theory. This domain is still in \emph{statu nascendi} and includes the following key research constructs: graph representation and reasoning; information extraction and graph construction; link data, information/knowledge integration and information/knowledge graph storage management; NLP understanding, semantic computing, and information/knowledge graph mining as well as information/knowledge graph application~\cite{graph-theory-2020}.
In the last decade, graphs in various social science and human life fields have become a source of innovative solutions~\cite{graph-theory-2020,Application-Graph-Theory-2021,Andreas-Luschow-2021}.

Interactions do not usually occur in an institutional vacuum; they are guided and constrained by agreed-on rules. It is also essential to understand the parameters that drive and constrain them. Hence, the design of institutional behaviors can be measured through Networks of Prescribed Interactions (NPIs), capturing patterns of interactions mandated by formal rules~\cite{Olivier2019}.
Social network analysis offers considerable potential for understanding relational data, which comprises  the contacts, ties, and 
connections that relate one agent to another and are not reduced to 
individual agent properties. 
In network analysis, relations 
express the linkages between agents. That analysis consists of a body of qualitative measures of network structure. Relational data is central to the principal concerns of social science tradition, emphasizing the investigation of the structure of social action. The structures are built from relations, and the structural concerns can be pursued through the collection and analysis of relational data~\cite{Social-Network-Analysis}.
Our approach uses social network analysis powered by institutional grammar.

\section{Our Prototype}
\label{sec:system}

\begin{figure}[!htb] %
    \centering
    \includegraphics[width=0.9\linewidth]{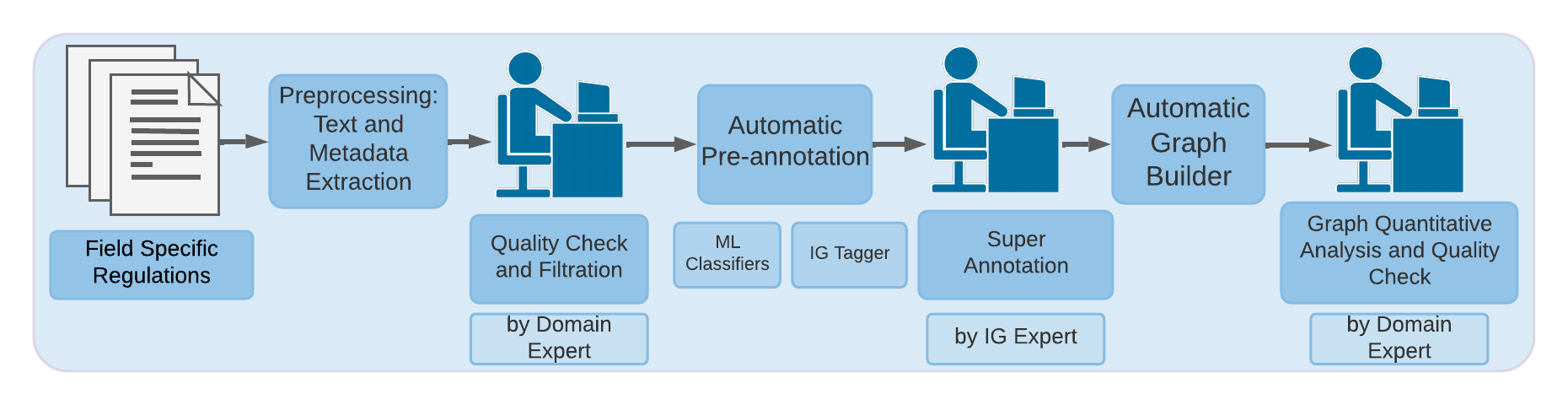}
    \caption{The analysis process of legal acts with our system support and manual work of IG human expert and refinement of a domain expert.}%
    \label{fig:IG_process}%
\end{figure}

Our prototype's workflow starts with setting up a crawler to collect legal documents from websites of interest. Its functionalities mirror the stages of work with legal acts: (1) retrieving documents from different internet resources and (2) selecting, for further analysis, only documents relevant for a specific IG-related research task (filtered by customizable defined rules or keywords). Then, (3) legal texts are prepared, pre-processed, and parsed with IG. The output -- selected IG components -- are then (4) refined and transformed into a graph where nodes represent institutional actors and objects, and the statements containing them are expressed as edges (relations). In this process, we also incorporate quality checks and super-annotations with an IG and domain expert in policy design. Finally, we consider institutional design research questions that can be answered using quantitative analysis based on the generated graph. Figure~\ref{fig:IG_process} shows this process of annotating legal documents by incorporating our automated tools in the prototype.
The first significant challenge regarding building the above system is the semantic understanding of legal policy texts. To automate the process of legal text analysis,  we utilize  Institutional Grammar as an intermediate layer for text interpretation and extraction of crucial entities. Thus, in our study, we incorporate IG as a text-semantic bridge. On the one hand, IG is a tool for describing and analyzing relations between institutional actors more systematically and structured than natural language texts. However, there are still open issues we are currently addressing with the manual work of policy design researchers. Firstly, there is still a need to define some IG statements more precisely, e.g., context data like Activation Conditions entities. Some IG entities are not defined precisely and are not structuralized as a graph structure or formal ontology with classes and logical rules between them. Such structurally ordered and precisely defined data (a graph/
an ontology) can further be used, for example, for defining rules of institutions and network analysis and reasoning~\cite{ig_social_network}. We have yet to manually address atomic statement extractions and classification (regulative vs. constitutive) and their transformations into active voice.

Our prototype's second challenge is transforming the IG format with the extracted entities into a graph structure and format, which is a proper data structure for automated analysis. Under the label of "social network analysis," social science has developed significant expertise in expressing information about particular elements of our reality in computer-readable data structures to analyze the interrelation between entities. This task is accomplished using IG as an interlayer between a legal text and mainly graph-based metrics expressing interrelations between entities.

\subsection{IG Tagger}

Institutional statements have two types -- regulative and constitutive -- with a few distinct IG tags. 
In IG, constitutive statements serve defining purposes, while regulative statements describe how actors' behaviors and actions should be regulated. For this task -- i.e., the differentiation of statement types -- we trained a classifier based on TF-IDF with 70 of the most relevant (1-3)-grams and Random Forest to distinguish between these types (our train set contained 249 statements). The model's F1-score is 92\% on the test dataset (84 observations out of 382).
\footnote{The model is used in the tagger's repository (\url{https://github.com/institutional-grammar-pl/ig-tagger}) and the experiments are included in a separate repository: \url{https://github.com/institutional-grammar-pl/sentence-types}.}

Then in our IG tagger, the first tagging stage, each word in the statement gets an annotation containing a lemma, a part of speech tag, morphological features, and relation to other words (extracted with Stanza package~\cite{qi2020stanza}). Due to different IG tags in regulative and constitutive statements, the automatic tagger has two different algorithms based on rules dedicated to each type of statement.
In the following, we show an example of tagger usage. For this purpose, we present only the selected rules in the analyzed sentence below. 
\begin{enumerate}
    \item \label{item:root_verb} If a sentence contains one word with \textit{root} tag and this word is a verb or an adjective:
    \begin{enumerate}
        \item \label{item:root_verb_function} If the word founded in~\ref{item:root_verb} is a verb, then annotate it as \textit{constitutive function}
        , otherwise as \textit{constituing properties}.
        \item \label{item:root_verb_function_2} If the word annotated in \ref{item:root_verb_function} has a child with \textit{aux:pass} or \textit{cop} relation, then annotate this child as \textit{constitutive function}.
         \item \label{item:entity} If the word annotated in \ref{item:root_verb_function} has a child with one of \textit{nsubj}, \textit{nsubj:pass} or \textit{expl} relation, then annotate this child as \textit{constituted entity}.
        \item \label{item:entity_4} If the word annotated in \ref{item:entity} has a child with one of \textit{det}, \textit{compound}, \textit{mark}, then annotate this child and all child's descendant as \textit{constituted entity}.
        
        \item \label{item:context} If the word annotated in \ref{item:root_verb_function} has a child with one of \textit{obl}, \textit{advmod}, \textit{xcomp}
relation, then annotate this child and all child's descendants as \textit{context}.
    \end{enumerate}
    \item \label{item:modal} If the word with \textit{root} tag has a child with \textit{aux} relation and that child's lemma is one of "must", "should", "may", "might", "can", "could", "need", "ought", "shall", then annotate this word as \textit{modal}.
\end{enumerate}

\begin{figure}[h!]
\centering
\includegraphics[width=0.7\linewidth]{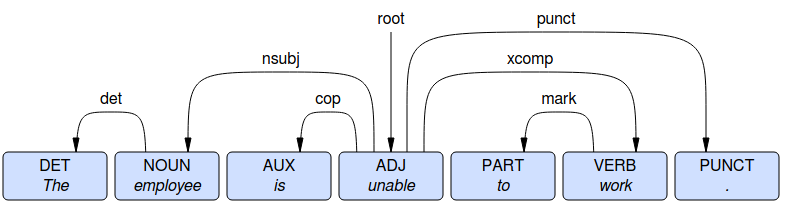}
\caption{Visualization of annotated constitutive sentence by Stanza NLP tagger in CONLL-U Viewer. 
}
\label{fig:conllu_viewer}
\end{figure}

Considering the sentence given in Figure~\ref{fig:conllu_viewer}, our tagging algorithm takes into account the following rules:
(1) According to rule~\ref{item:root_verb_function}, we annotate the word \textit{unable}  as \textit{constituting properties};
(2) According to rule~\ref{item:root_verb_function_2}, we annotate the word \textit{is} as \textit{constitutive function};  
(3) According to rules~\ref{item:entity} and~\ref{item:entity_4}, we annotate \textit{the employee} as \textit{constituted entity}; 
(4) According to rule~\ref{item:context}, we annotate words \textit{to work} as \textit{context}. 

Table~\ref{tab:eval_tager_detailed} shows overall tagger performance tested on the 2003 UNESCO Convention consisting of 142 regulative and 240 constitutive statements. For this analysis, predicted and correct tags were mapped before the evaluation: (A, prop) to (A), (B, prop) to (B) for regulative ones, and (E, prop) to (E), (P, prop) to (P) for constitutive statements (prop means property). Applied measures were determined based on the accuracy of the classification of the individual words in a sentence. 
The best results are achieved in recognizing Aim, Deontic, Function, and Modal components -- F1-score of over $99\%$. These tags are precisely defined. As we can see in Table~\ref{tab:eval_tager_detailed}, the constitutive statements and some tags of regulative statements are still a big challenge that should be defined more precisely and solved with machine learning-based text modeling.

\begin{table}[H]
    \centering
    \caption{Results of IG Tagger on regulative and constitutive statements.}
    \label{tab:eval_tager_detailed}
    \begin{tabular}{c|c|c|c|c}
        \textbf{Layer}  &
        \textbf{Component}  & 
         \textbf{F1 score}  &
         \textbf{Precision} &
         \textbf{Recall} \\
        \hline
        \multirow{6}{*}{\rotatebox[origin=c]{90}{\parbox[c]{1.5cm}{\centering \textbf{Regulative}}}}
        & Attribute &  0.51 & 0.62 & 0.56 \\
        &Object  &  0.55 & 0.67 & 0.61 \\
        & Deontic & 1.0  & 0.96 & 0.98 \\
        &Aim  & 0.99 & 0.86 & 0.92 \\
        & Context &  0.53 & 0.73 & 0.62\\
        &\textbf{Overall} & 0.71 & 0.69 & 0.68 \\
        \hline
        \multirow{6}{*}{\rotatebox[origin=c]{90}{\parbox[c]{2cm}{\centering \textbf{Constitutive}}}}
        & Entity & 0.79 &0.64 &0.71  \\
        &Property &0.57  &0.78 &0.66 \\
        &Function & 0.82 &0.82 & 0.82\\
        &Modal &1.00 & 0.83& 0.90\\
        &Context & 0.09& 0.02& 0.03 \\
        &\textbf{Overall} & 0.58 & 0.57 & 0.56 \\
    \end{tabular}
\end{table}

\subsection{Graph of Entities Extraction}

We extracted hypergraph information from our analyzed documents to get insights into a given document. The hypergraph vertices are the essential entities -- actors (e.g., institutions) and objects (e.g., legal documents). The statements (in the analyzed documents) represent connections (edges) in this structure. 

Let $H=(V, E)$ be a hypergraph, where verticles $V$ represent actors and objects appearing in the document. $E=(e_{1}, ..., e_{k}), e_{i} \subset V \forall_{i \in 1, ..., k}$ means hyperedges that are created where objects appear together in one statement from the analyzed legal documents. We chose hypergraphs, not graphs, because relations are formed by more than one vertex. 

After constructing the hypergraph, our analysis employed various graph measures, including the 
hypergraph centrality proposed in~\cite{scentrality}. We can also use other graph metrics to help describe information and relationships between entities and define the interrelations and expositions of particular actors.


\section{Use Case and Impact}
\label{sec:use_case}

Our use case is the 2003 UNESCO Convention, which establishes UNESCO listing mechanisms for the safeguarding of intangible cultural heritage, including the Representative List of the Intangible Cultural Heritage of Humanity, the List of Intangible Cultural Heritage in Need of Urgent Safeguarding, and the Register of Good Practices in Safeguarding of the Intangible Cultural Heritage~\cite{Blake2020,Schreiber2017}. 

As mentioned previously (Section~\ref{sec:intro}), we aim to study the institutional design of an international convention by comparing the actors' positions in the network of rules described by the convention. Following other researchers who transformed IG data into network data~\cite{Heikkila2018,Olivier2019,MESDAGHI2022120}, we use network metrics to describe the locations of actors in the institutional settings produced by the 2003 UNESCO Convention. We outline the Convention’s institutional design using data from its text explored using IG and network analysis. 

In addition, we compare metrics from network analysis with metrics 
from IG data. The comparison is the first step in understanding if an actor's position in a set of statements is a good indicator of their location in the institutional setting expressed as a graph. Looking at it from another angle, we aim to empirically verify to what extent indicators derived from IG analysis are good predictors of indicators related to network analysis. We want to determine whether future studies must process IG-derived data into data suitable for network analysis.

In our study, the position is considered from the perspective of two measures – visibility and centrality – referring to two sets of data: IG data and IG data converted into network data.  \textit{Visibility} of an actor is defined as the level of the directness with which institutional statements regulate the actor's actions. In the context of IG, this relates to the institutional statement components in which the actor is mentioned.  It can be measured on an ordinal scale built around the actor's place in a statement (see Table~\ref{tab:directness}). An actor mentioned in the most prominent elements of a statement (Attribute or Direct Object) has a higher score than an actor mentioned in concealed parts of statements (for example, properties of Objects). 
\begin{table}[H]
    \centering
    \caption{Actors' visibility scale in constitutive (CS) and regulative (RS) texts.}
    \label{tab:directness}
    \begin{tabular}{p{9cm}|c}
        \textbf{Actor:}  &
        \textbf{Weight} \\
        \hline        
        As Attribute in RS or Constituted Property in CS   & 6\\
        As Direct Object in RS or Constituting Entity in CS  & 5\\
        As Indirect Object in RS & 4\\
        In properties of Attribute or Constituted Property & 3\\
        In properties of  Direct Object or Constituting Entity & 2\\
        In properties of Indirect Object  & 1 \\
    \end{tabular}
\end{table}
Actors are ranked by their measures of directness. Each actor is assigned a weight depending on the class (see the ranking in Table~\ref{tab:directness}). Thus, we can define a \textit{visibility} measure as: 
$$ visibility = \sum_{c \in \{1, ..., C\}} w_{c} * \frac{n_{c}}{N},$$
where $N$ - number of statements, $n_{c}$ - occurrence in class $c$, $w_{c}$ - rank of class $c$.

\textit{Centrality} is associated with actors' positions in a network of statements. This analysis assumes an actor can be mentioned in less prominent IG components. However, at the same time, it could have a central role in the formal institutional setting created by a legal document. It is “involved” in many statements but is usually mentioned in subsidiary IG elements. Each statement is seen as an edge, and actors are depicted as nodes independent of their association with IG components. A set of all edges forms a hypergraph mapping the regulation. The measure of centrality in the hypergraph is computed for each actor. Then, the actors are ranked by their measures of centrality. 

Based on the above operationalizations, the research question that can be formulated for the exemplary case is: What is the relation of visibility versus centrality in the case of the Secretariat?
\begin{figure}[H]
\centering
\begin{subfigure}{.45\textwidth}
  \centering
  \includegraphics[width=0.9\linewidth]{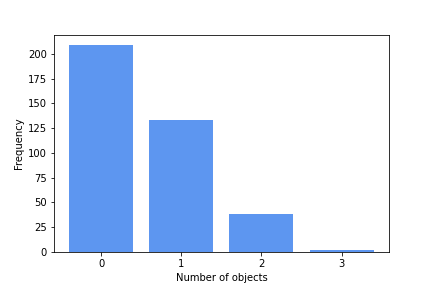}
  \caption{Actors.}
  \label{fig:sub1}
\end{subfigure}%
\begin{subfigure}{.45\textwidth}
  \centering
  \includegraphics[width=0.9\linewidth]{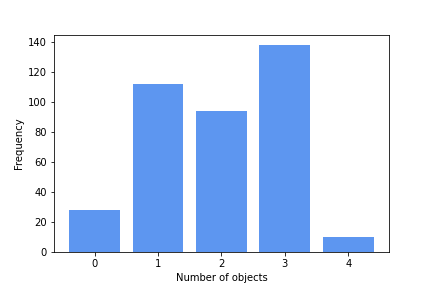}
  \caption{Actors and objects.}
  \label{fig:sub2}
\end{subfigure}
\caption{Histogram of mentions of actors and objects in statements which refer to the number of connected nodes in the hypergraph (hyperedges).}
\label{fig:hist_obj_occurance}
\end{figure}

\begin{figure}[H]
\centering
  \includegraphics[width=0.8\linewidth]{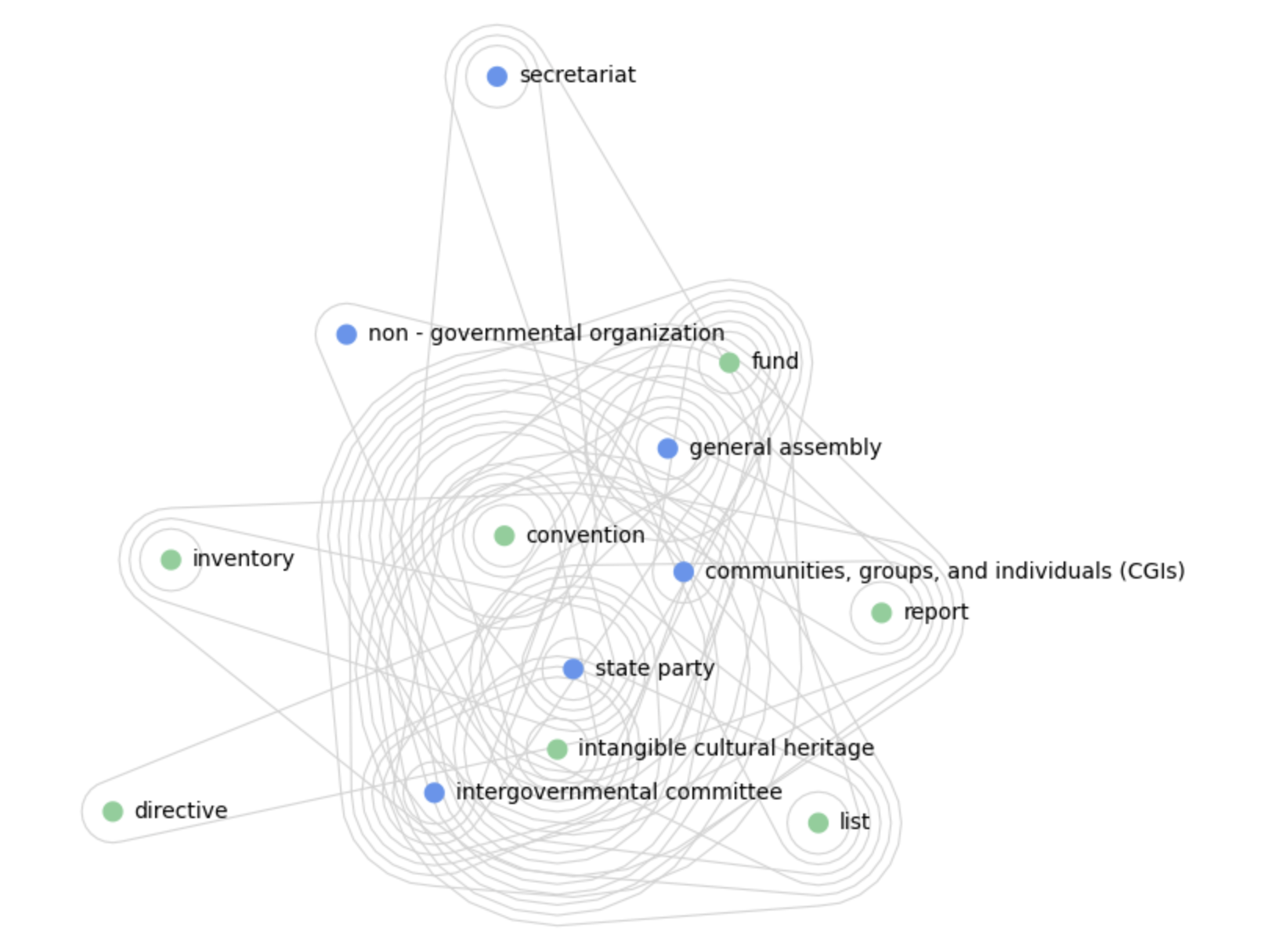}
 \caption{Actors and objects hypergraph 
 from the 2003 UNESCO Convention. Note: green dots denote objects, blue -- actors.
 }
\label{fig:graph}
\end{figure}

\begin{figure}[!h] 
  \centering
  \includegraphics[width=0.5\textwidth]{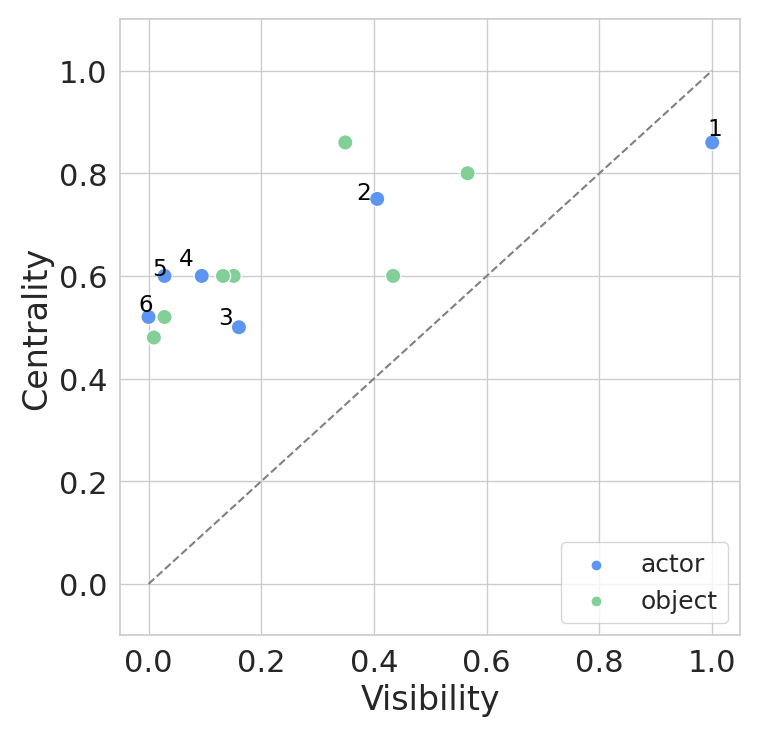}
\caption{Visibility vs Centrality (closeness centrality), where (1) is State Party, (2) -- General Assembly, (3) -- Secretariat, (4) -- Intergovernmental Committee, (5) -- Communities, Groups, and Individuals (CGIs), and (6) is Non-governmental organization. Note: green dots denote objects, blue -- actors.}
\label{fig:metrics-plot}
\end{figure}

Figure~\ref{fig:hist_obj_occurance} presents how many actors occur in one statement. The baseline centrality approach analyzes the occurrence of actors together within one sentence. As shown in Figure~\ref{fig:sub1}, usually, only one actor appears within one atomic sentence. Therefore, the analysis was expanded to consider the presence of actors and objects such as a report, individual or group. Figure~\ref{fig:sub2} illustrates that this approach has much greater potential. Then, finally, Figure~\ref{fig:graph} presents our hypergraph extracted from the 2003 UNESCO Convention.
Based on our analysis, we can answer the research question stating that the Secretariat belongs to a group of actors with low levels of visibility but relatively high normalized values of their centrality -- see  Figure~\ref{fig:metrics-plot}. This group comprises non-governmental organizations, the Intergovernmental Committee, Communities, Groups, and Individuals (CGIs), and the Secretariat. However, the Secretariat has the highest level of visibility in the group. It appears in more essential IG elements than the rest of the group. This observation indicates that the Secretariat's position in the institutional design of the 2003 UNESCO Convention is more important than a brief analysis of the Convention’s text reveals. However, the position of the Secretariat is not as undervalued from the perspective of visibility as the position of others in the group. The issue of undervaluation leads us to the second observation –- State Party is the only overexposed actor in the 2003 Convention. 

State Party is the only actor with a visibility score higher than its centrality score. However, we can only talk about a relative "overexposure" of the State Party -- it has, after all, the highest centrality score among all actors. We can tell that states that are parties to the Convention are crucial actors in the formal institutional design of the 2003 UNESCO Convention. This observation probably reflects the fact that international conventions are usually formulated in a manner that, above all, regulates the actions of actors being parties to international conventions -- the states themselves.

The third observation from our case study relates to the relationship between pure IG analysis and network analysis based on data provided by IG. The indicators coming from raw IG are poor proxies for network analysis ones. This observation shall encourage building systems that treat IG only as an interlayer for developing data on a formal institutional design from legal texts. 

\begin{table}[!h]
    \centering
    \caption{Visibility and centrality of actors and objects appearing in the document.}
    \label{tab:visibility}
    \begin{tabular}{lp{2.2cm}p{3.5cm}} 
        \textbf{Actor/ object}  &
        \textbf{Visibility} &
        \textbf{Closeness Centrality} \\
        \hline
        (1) State Party
        & 1.07 & 0.86\\
        Convention & 0.61 & 0.80\\
        Fund & 0.47 &  0.60\\
        (2) General Assembly & 0.44 &   0.75 \\
        Intangible cultural heritage & 0.38&  0.81\\
        (3) Secretariat  & 0.18 &  0.50 \\
        List & 0.17 & 0.60\\
        Report & 0.15 & 0.60\\
        Inventory & 0.04 &         0.52\\
        (4) Intergovernmental Committee & 0.11 &    0.6\\
        Directive & 0.02 &      0.48\\
        (5) Communities, groups, and individuals (CGIs) & 0.04 & 0.60\\
        (6) Non - governmental organization &    0.01 & 0.52 \\
    \end{tabular}
\end{table}

\FloatBarrier

\section{Conclusions}
\label{sec:conclusions}

This paper addresses the challenge of easily comprehending international agreements to solve global issues. It presents a prototype tool for efficiently extracting formal institutional design -- norms, rules, and actors -- from collected legal documents. The tool employs Institutional Grammar as a bridge between legal text and a graph representation of the institutional design, enabling practical NLP analysis of a complex legal text. 
The tool was tested in the 2003 UNESCO Convention. Its analysis reveals the position of the Convention Secretariat and other actors within the institutional design using the measures of visibility and centrality. The results indicate that the Secretariat holds average importance and is part of a group of relatively hidden actors in the Convention’s institutional design. 

The developed approach shows an analytical and research potential for optimizing and effectively designing the processes of formalizing any new organizations. The study demonstrates the potential of combining NLP tools, Institutional Grammar, and graph analysis to facilitate a deeper understanding of formal institutional design in international agreements. The resulting graph representation provides a clear, easily-comprehensible representation of the institutional structure and holds promise for advancing the field of computational diplomacy. The presented study also proves that a truly interdisciplinary approach to related areas of computational science and diplomacy can bring relevant and credible research results from theoretical and practical perspectives.   

\section{Acknowledgments}
The research was funded by the Centre for Priority Research Area Artificial Intelligence and Robotics of Warsaw University of Technology within the Excellence Initiative: Research University (IDUB) program (grant no 1820/27/Z01/\\ POB2/2021), by the Faculty of Mathematics and Information Science, and also by the European Union under the Horizon Europe grant OMINO (grant no 101086321).\footnote{Views and opinions expressed are however those of the author(s) only and do not necessarily reflect those of the European Union or the European Research Executive Agency. Neither the European Union nor European Research Executive Agency can be held responsible for them.}

Hanna Schreiber and Bartosz Pieliński wish to acknowledge that their contribution to this paper was carried out within the framework of the research grant Sonata 15, "Between the heritage of the world and the heritage of humanity: researching international heritage regimes through the prism of Elinor Ostrom's IAD framework," 2019/35/D/HS5/04247 financed by the National Science Centre (Poland).  

We want to thank many Students for their work under the guidance of Anna Wróblewska and Bartosz Pieliński on Institutional Grammar taggers and preliminary ideas of the system (e.g.~\cite{wichrowska_system_2021}), which we modified and extended further. We also thank the anonymous reviewers and the program committee for their helpful comments.


\FloatBarrier


\end{document}